## 9.8. APPENDIX 8.  Conference „Human language technologies 2005"proceedings

# MULTILINGUAL LEXICON DESIGN TOOL AND DATABASE MANAGEMENT SYSTEM FOR MT

**Gintaras Barisevicius Department of Software Engineering, Kaunas University of Technology (KTU), Lithuania**
**Bronius Tamulynas, Computer Networking Dept., KTU, Lithuania**

**Abstract**

The paper presents the design and development of English-Lithuanian-English dictionary-lexicon tool and lexicon database management system for MT. The system is oriented to support two main requirements: to be open to the user and to describe much more attributes of speech parts as a regular dictionary that are required for the MT. Programming language Java and database management system MySql is used to implement the designing tool and lexicon database respectively. This solution allows easily deploying this system in the Internet. The system is able to run on various OS such as: Windows, Linux, Mac and other OS where Java Virtual Machine is supported. Since the modern lexicon database managing system is used, it is not a problem accessing the same database for several users.

## 1. Introduction

The design and development of English-Lithuanian-English (ELE) dictionary-lexicon tool and lexicon database management system (DMS) for MT is oriented to support two main requirements: to be open to the user and to describe much more attributes of speech parts as regular dictionary that are required for the MT. Currently, the lexicon is designed to support all parts of speech for Lithuanian and English languages. It is possible to extend the same database for other pairs of languages as well. Polysemy problem is overcome adding an additional table between two tables linking different translations of the word in the target language. The translations for the same words are enumerated in descending priority in both directions. In this way it is possible to ensure that even if the translation won't be very exact, the user will be able to choose the suitable words himself. The possibility to include additional context attributes for the nouns is allowed. Thus the meaning of the word to be translated first will be chose according to the context and highest priority. The user friendly interface is quite comfortable in that way that it is possible to see all generated forms and that is very efficient in the process of filling dictionary with new words. There is ambition to use some additional interface features that might improve user work with MT system:  such as other meaning choice or selection of words that are being translated and etc. So, if the translation won't be very exact, the user will be able to define the suitable words adequacy himself. Java and database management system MySql is used to implement the design tool and lexicon database respectively. This solution allows easily to be deployed the MT system in the Internet. The system is available to run on various OS such as: Windows, Linux, Mac and other OS where Java VM is supported. Currently, only one lexicographer can work on the system simultaneously, but the import (merging two dictionaries into one) function is under way. Since we are using modern DMS, it is not a problem accessing the same database for several users and mutual exclusion principle is handled by database managing system itself. The Lithuanian Government approved to support this project according to the national program "Lithuanian language in Information society for the years 2005-2006 for the development of the Lithuanian language technologies including computer-based translation".

**2. The linguistic Databases and Electronic Dictionaries for MT: problems and solutions**

For a long time there was only one bidirectional electronic dictionary WinLED in Lithuania (VteX company http://www.led.lt/) for Lithuanian-English and Lithuanian-German language pairs. The dictionary was comfortable to use, since it had a „copy and translate" function, which allows automatically translate the text copied to the OS clipboard. Nevertheless, WinLED has several disadvantages: the user interface is in English only, the wordlist is too short and misleading translation can be occurring.

Just recently a new dictionary „Tildės biuras" has been distributed *(http://www.tilde.lt/ biuras).* It also maintain bidirectional Lithuanian-English and Lithuanian-German dictionaries. English-Lithuanian dictionary contains 50000 words and phrases. The dictionary proposes not only the translations, but word pronunciation, part of speech and usage of the word. The Lithuanian user interface is quite important since a major part of Lithuanian population is barred from e-content due the lack of good English skills.

An electronic version of B.Piesarskas „Didysis anglų-lietuvių kalbos žodynas" (Great dictionary of English-Lithuanian languages) „Alkonas" (*http://www.fotonija.lt)* is more comfortable than previous dictionaries, since it contains the book format and has a huge word list (around 100000) with the words usage examples. The translation is bidirectional, but the problem is that words are indexed only from English to Lithuanian language and Lithuanian word translation search is performed using the same index.

All these dictionaries are only alternatives for the paper dictionaries since they are more comfortable and are possessed by faster search engine. Nevertheless, they have no additional data that is required for the MT, such as declensions, conjugations and etc. These available electronic dictionaries are generally different from those that are needed for the MT system. The problem is concerned with the requirement to perform not only word-to-word translation but to adjust the syntactic and semantic information as well as further processing of the target text.

The 100 million word text corpora for Lithuanian language (http://donelaitis.vdu.lt) is a very important resource for building new lexicons for MT. There are still no operational parallel corpora for English and Lithuanian languages. Naturally, the building such corpora is time and effort consuming, since its designing process is rather complex. As an alternative for parallel corpora could be comparable corpora. Such corpora should contain texts in both English and Lithuanian languages and disseminated according to the topics of the texts.

The research on Computer-Based Translation (CBT) was initiated at KTU in 2001 (Tamulynas 2004: 16–19). Several prototypes of CBT systems have been produced, but every time the same problems have been encountered: they were robust and incomplete in terms of lexical data (Misevičius et al. 2002: 38–45).

**3. Conceptual linguistic structure of lexicological database model**

Lexicon for MT (Hutchins 1992) or (Trujillo 1999) is a main and most important part of the system. Nevertheless there is no theoretical background what conceptual structure of lexicological database model should prevail (Melamed 1998). The fundamental requirements connected with the functionality of CBT system are: flexibility, correctness, translation quality and etc.

According to the Lithuanian grammar (Ambrazas et al 1997) it seems enough to keep the word stem in the dictionary only. Actually, this information is not sufficient. We must keep the conjugations, genders, declension, numbers and other attributes. Suitable forms of the Lithuanian words can be generated according to the Lithuanian grammar rules. The stem then is concatenated with the endings (sometimes the stem also has to be modified).

Prepositions take a very important role in the text. They usually determine the case of the word in the text. So, they must be stored with reference what case do they require. The translation of preposition combinations (phrases) will be incorporated into the phrase translation. The conceptual structure of ELE dictionary (lexicon) consists of:

- English words with grammatical and semantic attributes as well as with additional references to the roles they can perform in the sentence, when they belong to one or another part of speech, e.g. the noun can be object, subject and etc.
- corresponding Lithuanian words with all grammatical and semantic attributes.

The semantic information about the correspondence in English and Lithuanian languages should also be included and possible translation variations might be interpreted. The semantic information consists of the rules or prescriptions how the words may be used in certain context and etc.

Previous CBT prototypes (Misevičius et al. 2002: 38–45) didn't support all parts of speech and was lacking some morphological forms of those parts of speech that have been implemented. Noteworthy, that this new ELE version overtakes all English and Lithuanian parts of speech including noun, verb, adjective, pronoun, numeral etc. A large number of all generated morphological forms (see Table 12) are displayed in tables. In this case it is very convenient to compare the complexity, variety and to check differences between languages. For example, the variety (Ambrazas et al 1997) of Lithuanian regular noun has 14 forms, adjective 147 forms and verb over than 229 (around 400). To enter all those words manually, would be time and effort consuming, so automating this process as much as possible is a fair solution. The ending is the variable part of word, which shows the relation with other words in the sentence.

**Table 12 Some of Grammatical Categories of Lithuanian and English**

| Part of Speech | English | Lithuanian |
|---|---|---|
| Noun | • Feminine and masculine genders.<br>• Number: the singular, the plural; singular nouns (*milk*), plural nouns (*scissors*).<br>• The genitive case (*boy – boy's*). | • Feminine and masculine genders.<br>• Number: the singular, the plural; singular nouns (*laimė*), plural nouns (*žirklės, akiniai*).<br>• Case is of great importance: it indicates a relationship of a word with other words in a collocation and the sentence. There are 6 cases.<br>• There are 5 declensions. |
| Verb | • Regular and irregular verbs<br>• Conjugating an auxiliary verb+ standard form of the verb<br>• Main categories of tenses are present, past and future tenses (there are 12 tenses overall)<br>• Modal and auxiliary verbs. | • Inflective by person.<br>• Possesses the category of mood (direct, subjunctive, imperative).<br>• 4 tenses: present, 2 past, future.<br>• Characterized by number.<br>• Forms, non-inflective by person: the infinitive, participles. |

The meaning of word is derived according to its possible usage, relation with other parts of speech and its place in the sentence, context and etc. The predicate (verb) is most important since it determines how other words are situated in the sentence and how can they be linked (context dependency).

The structure of the dictionary is oriented to the objects. Thus ELE lexicon DB is open and operates virtually, i.e. there is a possibility to add new words, terms, phrases or expressions. This property greatly increases the translation quality. Word dissemination to notional categories makes sentence analysis easier in case the word has several possible translations. It helps to choose one translation according to the meaning. On numerous occasions correct sentence translation is possible when the particular word meaning is determined by the grammar rules. The polysemy is realized to both translation directions using an additional table which links two tables of different parts of speech.

Since the system is planned to be developed further, it is desirable to have an additional interface features that might improve user case with CBT. The user interface should be able to help the user to choose the translation alternatives for the possible translation list, as well as to use the history of translation.

**4. Multilingual Lexicon design and database management tools**

We have chosen very flexible way of implementing our lexicon (Barisevičius, Černys 2004). Since we define our queries to the data using SQL (Standard Query Language), so the exact DMS is not essential (we use MySql). The usage of DMS make the lexicon easy to modify: to add new attributes, delete them, or modify the names or types and etc. It is possible to extend the same database to other human languages as well. Since MySql can store up to 2Gb of data we don't have to worry about the volumes anymore. The running test with database of 20 million words and the retrieval of the first word took 0.03 seconds, which was longer due to making the connection to the database. The retrieval time of the following searches was less than 0.01 seconds.

We considered the domain possibility for the nouns, thus user can choose the priorities of finding words in certain domain. This way user can prioritize the word translation according to his domain of interest.

While Java programming language for implementation is used it is possible to make system available on-line. Naturally, there is a small disadvantage of Java that it is quite slow, but the recent releases of Java Runtime Environment had improved their performance a lot. Shrinking the compiled code is another option to increase the performance. As the names of variables are shrunk to be minimal, the system operates with shorter names. Besides, the system is available on Windows, Linux, Mac and other OS where the Java Virtual Machine is supported.

The user friendly interface is designed to optimize the work of the lexicographer, that he could select the attributes of the word and generate all possible morphological forms. This feature should save amounts of lexicographers work, since he won't have to enter each form individually and besides will be able to see all generated forms in the screen in a very compact way.

Another problem of filling is for the lexicographers to work simultaneously. This problem can be easily solved if the system is running on-line. Since, we use state-of-art DMS, we can access the data simultaneously. All the mutual exclusion problems are handled by the database system and we don't have to worry about that. The database export and import functions are under way. Till the end of the year, we are planning to fill the dictionary with around 20,000 of words. That should be enough for primitive translation using the most frequent words in the language.

**5. Conclusions**

English-Lithuanian-English dictionary-lexicon tool and lexicon database management system for MT is oriented to support two main requirements: to be open to the user and to describe much more attributes of speech parts as a regular dictionary that are required for the MT. Programming language Java and database management system MySql is used to implement the designing tool and lexicon database respectively. This solution allows easily deploying this system in the Internet. The system is able to run on various OS where Java Virtual Machine is supported. Currently, only one lexicographer can work on the system simultaneously, but the import (merging two dictionaries into one) function is under way. The research is performed according to the national program "Lithuanian language in Information society for the years 2005-2006 for the development of the Lithuanian language technologies including computer-based translation". The Tool was demonstrated in DWS (Dictionary Writing Systems workshop, Brno) in September 6-7, 2004, as well as in workshop took place in Vytautas Magnus University in October 26, 2004.